\documentclass[mlabstract]{jmlr}

\usepackage{longtable}%

\usepackage{booktabs}
\usepackage[load-configurations=version-1]{siunitx} %
\theorembodyfont{\upshape}
\theoremheaderfont{\scshape}
\theorempostheader{:}
\theoremsep{\newline}

\jmlrvolume{}
\firstpageno{1}

\jmlryear{2024}
\jmlrworkshop{Symmetry and Geometry in Neural Representations}

\title[Neural Network Symmetrisation in Concrete Settings]{Neural Network Symmetrisation in Concrete Settings}

  \author{\Name{Rob Cornish} \Email{rob.cornish@stats.ox.ac.uk} \\ \addr Department of Statistics, University of Oxford}

\newcommand{\R}{\mathbb{R}}

\newcommand{\C}{\mathsf{C}}

\newcommand{\Set}{\mathsf{Set}}

\newcommand{\Stoch}{\mathsf{Stoch}}

\newcommand{\id}{\mathrm{id}}

\newcommand{\sym}{\mathsf{sym}}
\newcommand{\step}{\mathsf{func}}
\newcommand{\ave}{\mathsf{ave}}

\usepackage{marginnote}

\usepackage{bm}
\usepackage{mathtools}
\usepackage{tikz-cd}
\usepackage{tikzit}

\tikzstyle{morphism}=[fill=white, draw=black, shape=rectangle]
\tikzstyle{medium box}=[fill=white, draw=black, shape=rectangle, minimum height=0.8cm]
\tikzstyle{medium large morphism}=[fill=white, draw=black, shape=rectangle, minimum height=1.2cm]
\tikzstyle{large morphism}=[fill=white, draw=black, shape=rectangle, minimum height=1.7cm]
\tikzstyle{bn}=[fill=black, draw=black, shape=circle, inner sep=1.5pt]
\tikzstyle{state}=[fill=white, draw=black, regular polygon, regular polygon sides=3, minimum width=0.8cm, shape border rotate=180, inner sep=0pt]
\tikzstyle{long state}=[fill=white, draw=black, shape=isosceles triangle, isosceles triangle apex angle=90, shape border rotate=270]
\tikzstyle{medium state}=[fill=white, draw=black, regular polygon, regular polygon sides=3, minimum width=1.3cm, inner sep=0pt, shape border rotate=180]
\tikzstyle{large state}=[fill=white, draw=black, regular polygon, regular polygon sides=3, minimum width=2.2cm, shape border rotate=180, inner sep=0pt]
\tikzstyle{wn}=[fill=white, draw=black, shape=circle, inner sep=1.5pt]
\tikzstyle{likelihood}=[fill=white, draw=black, regular polygon, regular polygon sides=3, minimum width=0.8cm, shape border rotate=0, inner sep=0pt]

\tikzstyle{arrow}=[->]
\tikzstyle{dashed box}=[-, dashed]
\tikzstyle{new edge style 0}=[-, fill={rgb,255: red,148; green,162; blue,255}, draw=none]

\begin{document}

\maketitle

\vspace{-2em}
\begin{abstract}
\cite{cornish2024stochastic} recently gave a general theory of neural network \emph{symmetrisation} in the abstract context of Markov categories.
We give a high-level overview of these results, and their concrete implications for the symmetrisation of deterministic functions and of Markov kernels.
\end{abstract}

\section{Introduction}
\label{sec:intro}

It is often useful to ensure that a neural network $f : X \to Y$ is \emph{equivariant} with respect to the actions of some group.
Recently there has been interest in doing so via \emph{symmetrisation} techniques \citep{yarotsky2018universal}.
Roughly speaking, these approaches take $f \coloneqq \sym(f_0)$ where $f_0 : X \to Y$ is some neural network that is \emph{not} equivariant, and $\sym$ is a function that maps non-equivariant neural networks to equivariant ones.

A variety of different choices of $\sym$ have been considered in the literature to-date, including Janossy pooling \citep{murphy2018janossy}, frame averaging \citep{puny2022frame}, canonicalisation \citep{kaba2023equivariance}, and probabilistic averaging \citep{jinwoo2023learning,dym2024equivariant}.
Recently, \cite{cornish2024stochastic} gave a general theory of symmetrisation that characterises all possible choices of $\sym$, recovering these previous techniques as special cases, and extending to the novel setting of \emph{stochastic} neural networks $f$, which had not previously been considered.
This framework also streamlines the presentation of compositional and recursive usage of these techniques, and encompasses a range of complex situations such as noncompact translation groups and semidirect products in a uniform way.

The results of \cite{cornish2024stochastic} were developed in terms of \emph{Markov categories} \citep{fritz2020synthetic}, which provide a high-level algebraic framework for reasoning about probability in an intuitive yet rigorous way.
However, Markov categories are currently not widely known in the machine learning community, and so in this paper we present special cases of \cite{cornish2024stochastic} in more familiar concrete settings.
We begin in Section \ref{sec:deterministic-symmetrisation} with the purely deterministic case considered by \cite{kaba2023equivariance}, and then extend to include randomness in Section \ref{sec:stochastic-symmetrisation}.
We assume knowledge of only the basic definition of a category (see Appendix \ref{apd:basic-category-theory}).

\paragraph{Notation}

Given a category $\C$, we will denote by $\C(X, Y)$ the set of morphisms $X \to Y$ in $\C$.
We also denote the category of sets and functions by $\Set$.

\section{Deterministic symmetrisation} \label{sec:deterministic-symmetrisation}

\paragraph{Actions}

Recall that an \emph{action} of a group $G$ on a set $X$ is a function $\alpha_X$ that takes as input some $g \in G$ and $x \in X$ and returns an output $\alpha_X(g, x)$ that also lives in $X$.
We will usually write this output simply as $g \cdot x$ when $\alpha_X$ is clear from context.

\paragraph{The category of $G$-sets}

For every group $G$, there is a category $\Set^G$.
Each object of this category is a \emph{$G$-set}, or in other words a set $X$ equipped with a group action $\alpha_X$.
The morphisms $f : X \to Y$ in $\Set^G$ are then functions that are \emph{equivariant} with respect to $\alpha_X$ and $\alpha_Y$, so that
\[
  f(g \cdot x) = g \cdot f(x)
\]
holds for all $x \in X$ and $g \in G$.
\emph{Invariant} functions are the special case where $\alpha_Y$ is the trivial action with no effect, in which case we have $f(g \cdot x) = f(x)$.

Given two $G$-sets $X$ and $Y$, their \emph{product} is always another $G$-set that we denote by $X \otimes Y$.
This is simply the cartesian product of $X$ and $Y$ equipped with the \emph{diagonal} action $\alpha_{X \otimes Y}$ defined as $g \cdot (x, y) \coloneqq (g \cdot x, g \cdot y)$.

\paragraph{Symmetrisation procedures}

Suppose $H \subseteq G$ is a subgroup.
Given any $G$-set $X$, we can always obtain an $H$-set $R X$ with the same underlying set as $X$, and whose $H$-action is obtained via \emph{restriction}, so that $\alpha_{R X}(h, x) \coloneqq \alpha_X(h, x)$.
This allows us to define a \emph{symmetrisation procedure} as any function of the form:
\begin{equation} \label{eq:sym-proc-general-form}
  \begin{tikzcd}
    \Set^H(R X, R Y) \ar{r}{\sym} & \Set^G(X, Y).
  \end{tikzcd}
\end{equation}
Notice that such a $\sym$ takes a function that is equivariant only with respect to the subgroup $H$ and ``upgrades'' it to become equivariant with respect to the whole group $G$.
If $H$ is the trivial subgroup consisting of just the identity element, then $H$-equivariance becomes trivial, and $\sym$ may take as input any arbitrary function $X \to Y$.
The latter case has received the majority of attention in the symmetrisation literature, although Section 3.3 of \cite{kaba2023equivariance} considers the general setup here.

Although we focus on \eqref{eq:sym-proc-general-form}, Definition 4.7 of \cite{cornish2024stochastic} formulates symmetrisation slightly more generally in terms of a homomorphism $\varphi : H \to G$.
The case of \eqref{eq:sym-proc-general-form} is then recovered by considering a subgroup inclusion $\varphi : H \hookrightarrow G$.
As we explain in Appendix \ref{apd:sym-along-homomorphism}, this approach can be more convenient for describing the composition of multiple symmetrisation procedures in sequence, which allows for ``building up'' complex equivariance properties in a structured way.

\paragraph{A general characterisation}

The following result allows us to characterise all symmetrisation procedures, i.e.\ functions of the form \eqref{eq:sym-proc-general-form}.
Here $G/H$ denotes the set of $H$-\emph{cosets}
\[
  G/H \coloneqq \{[g] \mid g \in G\},
\]
where $[g] \coloneqq \{gh \mid h \in H\}$.
This set always comes equipped with the $G$-action $g \cdot [g'] \coloneqq [gg']$, and hence may always be regarded as an object in $\Set^G$.
We now have the following result.

\begin{theorem} \label{prop:relative-left-adjoint}
  For all choices of the various components involved, there is a bijection
  \begin{equation} \label{eq:relative-left-adjoint-3}
    \begin{tikzcd}%
      \Set^H(R X, R Y) \ar{r}{\cong} & \Set^G(G/H \otimes X, Y)
    \end{tikzcd}
  \end{equation}
  that sends $f \mapsto f^\sharp$ defined as $f^\sharp([g], x) \coloneqq g \cdot f(g^{-1} \cdot x)$.
\end{theorem}

We give a self-contained proof in Appendix \ref{apd:proof-of-theorem}, where we also explain how this result arises very naturally in the context of category theory.

It now follows that the following two steps always constitute a symmetrisation procedure for every choice of the function $\step$ shown:
\begin{equation} \label{eq:sym-characterisation}
   \begin{tikzcd}%
      \Set^H(R X, R Y) \ar{r}{\cong} & \Set^G(G/H \otimes X, Y) \ar{r}{\step} & \Set^G(X, Y).
    \end{tikzcd}
\end{equation}
Here the first arrow denotes \eqref{eq:relative-left-adjoint-3}.
Moreover, since \eqref{eq:relative-left-adjoint-3} is a bijection, \emph{every} symmetrisation procedure can be obtained in this way for some choice of $\step$.
In other words, \eqref{eq:sym-characterisation} fully characterises all possible functions of the desired form \eqref{eq:sym-proc-general-form}.

\paragraph{Precomposition}

The question now is, how can we obtain $\step$?
If we want a ``general purpose'' strategy that works without further assumptions, then there is only one obvious choice.
This is namely the mapping that sends $f^\sharp : G/H \otimes X \to Y$ in $\Set^G$ to the composition
\begin{equation} \label{eq:precomposition-1}
  \begin{tikzcd}%
    X \ar{r}{\Gamma} & G/H \otimes X \ar{r}{f^\sharp} & Y
  \end{tikzcd}
\end{equation}
where $\Gamma$ is any choice of morphism $X \to G/H \otimes X$ in $\Set^G$.
In other words,
\[
  \step(f^\sharp) \coloneqq f^\sharp \circ \Gamma.
\]
It follows directly from the fact that $\Set^G$ is a category and hence closed under composition that $\step$ here is a function of the type required in \eqref{eq:sym-characterisation}.

A reasonable condition for a symmetrisation procedure $\sym$ to satisfy is \emph{stability} \cite[Definition 4.11]{cornish2024stochastic}: it should be the case that, if $f$ is already $G$-equivariant, then
\[
  \sym(f) = f.
\]
When $\step$ is obtained via precomposition as in \eqref{eq:precomposition-1}, this holds if and only if $\Gamma : X \to G/H \otimes X$ can be written as
\[
  \Gamma(x) = (\gamma(x), x)
\]
for some $\gamma : X \to G/H$ in $\Set^G$ \cite[Proposition 5.5]{cornish2024stochastic}.
In other words, $\Gamma$ must have the following ``shape'' (read from left to right):
\begin{equation} \label{eq:precomposition-morphism}
  \begin{tikzpicture}
	\begin{pgfonlayer}{nodelayer}
		\node [style=none] (12) at (4.75, 1) {};
		\node [style=none] (13) at (6, 1) {};
		\node [style=none] (14) at (4.75, -1) {};
		\node [style=none] (15) at (6, -1) {};
		\node [style=none] (16) at (3, 0) {};
		\node [style=none] (17) at (3.75, 0) {};
		\node [style=none] (18) at (7, 1) {$G/H$};
		\node [style=none] (19) at (6.5, -1) {$X$};
		\node [style=none] (20) at (2.5, 0) {$X$};
		\node [style=bn] (21) at (3.75, 0) {};
		\node [style=morphism] (22) at (5.25, 1) {$\gamma$};
	\end{pgfonlayer}
	\begin{pgfonlayer}{edgelayer}
		\draw (14.center) to (15.center);
		\draw (12.center) to (13.center);
		\draw (17.center) to (16.center);
		\draw [bend right=315] (14.center) to (17.center);
		\draw [bend left=45] (17.center) to (12.center);
	\end{pgfonlayer}
\end{tikzpicture}

\end{equation}

\paragraph{End-to-end procedure}

With $\step$ obtained in this way, the overall symmetrisation procedure \eqref{eq:sym-characterisation} maps $f : R X \to R Y$ in $\Set^H$ to the following morphism $X \to Y$ in $\Set^G$:
\begin{equation} \label{eq:symmetrised-morphism}
  \begin{tikzpicture}
	\begin{pgfonlayer}{nodelayer}
		\node [style=none] (0) at (-0.75, 0.5) {};
		\node [style=none] (1) at (3.25, 0.5) {};
		\node [style=none] (2) at (-0.75, -1.5) {};
		\node [style=none] (3) at (10.5, -1.5) {};
		\node [style=none] (4) at (-2.5, -0.5) {};
		\node [style=none] (5) at (-1.75, -0.5) {};
		\node [style=none] (8) at (-3, -0.5) {$X$};
		\node [style=bn] (9) at (-1.75, -0.5) {};
		\node [style=morphism] (10) at (-0.25, 0.5) {$\gamma$};
		\node [style=morphism] (11) at (3.5, 0.5) {$s$};
		\node [style=none] (12) at (3.75, 0.5) {};
		\node [style=none] (13) at (5.25, 0.5) {};
		\node [style=none] (14) at (6.25, 1.5) {};
		\node [style=none] (15) at (6.25, -0.5) {};
		\node [style=bn] (16) at (5.25, 0.5) {};
		\node [style=medium box] (17) at (11, -1) {$\alpha_X$};
		\node [style=none] (18) at (10.5, -0.5) {};
		\node [style=morphism] (19) at (7.5, -0.5) {$(-)^{-1}$};
		\node [style=none] (20) at (11.25, -1) {};
		\node [style=none] (21) at (13.5, -1) {};
		\node [style=morphism] (22) at (13.75, -1) {$f$};
		\node [style=none] (23) at (14, -1) {};
		\node [style=none] (24) at (16.25, 1.5) {};
		\node [style=none] (25) at (16.25, -1) {};
		\node [style=large morphism] (26) at (16.5, 0.25) {$\alpha_Y$};
		\node [style=none] (27) at (16.75, 0.25) {};
		\node [style=none] (28) at (18.25, 0.25) {};
		\node [style=none] (29) at (18.75, 0.25) {$Y$};
		\node [style=none] (38) at (2.75, 2.25) {};
		\node [style=none] (39) at (2.75, -2.25) {};
		\node [style=none] (40) at (17.75, -2.25) {};
		\node [style=none] (41) at (17.75, 2.25) {};
		\node [style=none] (42) at (1.5, 1) {$G/H$};
		\node [style=none] (43) at (4.5, 1) {$G$};
		\node [style=none] (44) at (9.5, 0) {$G$};
		\node [style=none] (45) at (12.5, -0.5) {$X$};
		\node [style=none] (46) at (15, -0.5) {$Y$};
	\end{pgfonlayer}
	\begin{pgfonlayer}{edgelayer}
		\draw (2.center) to (3.center);
		\draw (0.center) to (1.center);
		\draw (5.center) to (4.center);
		\draw [bend right=315] (2.center) to (5.center);
		\draw [bend left=45] (5.center) to (0.center);
		\draw (13.center) to (12.center);
		\draw [bend left=45] (15.center) to (13.center);
		\draw [bend left=45] (13.center) to (14.center);
		\draw (18.center) to (15.center);
		\draw (21.center) to (20.center);
		\draw (24.center) to (14.center);
		\draw (23.center) to (25.center);
		\draw (28.center) to (27.center);
		\draw [style=dashed box] (40.center) to (39.center);
		\draw [style=dashed box] (39.center) to (38.center);
		\draw [style=dashed box] (41.center) to (38.center);
		\draw [style=dashed box] (40.center) to (41.center);
	\end{pgfonlayer}
\end{tikzpicture}

\end{equation}
Here the dashed box denotes $f^\sharp$ obtained from \eqref{eq:relative-left-adjoint-3}, with $s : G/H \to G$ any choice of \emph{coset representative}, so that $s([g]) \in [g]$.
Likewise, $(-)^{-1}$ just inverts its input, sending $g \mapsto g^{-1}$.
Note that \eqref{eq:precomposition-morphism} is simply ``plugged in'' to the dashed box as per the precomposition step \eqref{eq:precomposition-1}.

In more traditional notation, letting $h \coloneqq s \circ \gamma$, the result \eqref{eq:symmetrised-morphism} maps $x \in X$ to the value
\[
  h(x) \cdot f(h(x)^{-1} \cdot x),
\]
which recovers the canonicalisation architecture of \cite{kaba2023equivariance} (see their Theorem 3.1).
The account given here therefore provides a theoretical explanation of how this architecture arises.
Additionally, as we describe next, this same story can now be generalised beyond $\Set$ to incorporate \emph{stochasticity}, which is useful for many practical applications.

\section{Stochastic symmetrisation} \label{sec:stochastic-symmetrisation}

\paragraph{Markov kernels}

Every component considered above (e.g.\ $f$ and $\gamma$ in \eqref{eq:symmetrised-morphism}) was a deterministic function.
In this section, we now allow these to depend on some additional randomness.
We do so by formalising these components as \emph{Markov kernels} instead of functions.
We gloss over some mathematical details here, giving additional background in Appendix \ref{apd:markov-kernels} instead.

A Markov kernel $k : X \to Y$ models a \emph{conditional distribution} or \emph{stochastic function} that, when given an input $x \in X$, produces a random output in $Y$ with distribution $k(dy|x)$.
For most Markov kernels of interest, $k(dy|x)$ is obtained as the distribution of some random variable $f(x, \bm{U})$, where $\bm{U}$ is a random variable, and $f$ is a (deterministic) function.
By letting $\bm{U}$ be constant, every deterministic function $X \to Y$ can be regarded as a Markov kernel that happens to be deterministic.
Markov kernels can be composed and so form a category called $\Stoch$ (see Appendix \ref{apd:markov-kernels}).

\paragraph{Stochastic equivariance}

Suppose $G$ is a group acting on $X$ and $Y$.
It is natural to say that a Markov kernel $k : X \to Y$ is \emph{equivariant} if for all $g \in G$ and $x \in X$ we have
\begin{equation} \label{eq:stochastic-equivariance}
  k(dy|g \cdot x) = g \cdot k(dy|x),
\end{equation}
where the right-hand side denotes the \emph{pushforward} of $k(dy|x)$, i.e.\ the distribution of $g \cdot \bm{Y}$ when $\bm{Y} \sim k(dy|x)$.
When $k(dy|x)$ is given by the distribution of some $f(x, \bm{U})$ as described above, this condition says
\[
  f(g \cdot x, \bm{U}) \overset{\mathrm{d}}{=} g \cdot f(x, \bm{U}),
\]
where $\overset{\mathrm{d}}{=}$ is equality in distribution.
Likewise, if $k(dy|x)$ has a conditional \emph{density} $p(y|x)$, then Proposition 3.18 of \cite{cornish2024stochastic} shows that \eqref{eq:stochastic-equivariance} also follows from the notion of stochastic equivariance commonly seen in the machine learning literature, which says that
\begin{equation} \label{eq:stochastic-equivariance-density}
  p(g \cdot y|g \cdot x) = p(y|x)
\end{equation}
always holds (see e.g.\ \citep{xu2022geodiff,hoogeboom2022equivariant}).
Note however that \eqref{eq:stochastic-equivariance-density} implicitly assumes the action on $Y$ has unit Jacobian: if this is not the case (such as for scale transformations), a Jacobian term should appear on the left-hand side instead.
In contrast, the condition \eqref{eq:stochastic-equivariance} always makes sense regardless of this.

\paragraph{Stochastic symmetrisation}

We now consider how to define symmetrisation procedures for Markov kernels.
The key point of \cite{cornish2024stochastic} is that the developments in Section \ref{sec:deterministic-symmetrisation} can be generalised beyond $\Set$ to any \emph{Markov category}.
It turns out that $\Stoch$ is also a Markov category \cite[Section 4]{fritz2020synthetic}, which gives rise to a framework for symmetrising Markov kernels directly as a special case.
We sketch this now.

Like in $\Set$, every group $G$ gives rise to a category $\Stoch^G$ whose morphisms are $G$-equivariant Markov kernels.
Given a subgroup $H \subseteq G$, we can define a \emph{symmetrisation procedure} just as in \eqref{eq:sym-proc-general-form}, with $\Set$ replaced by $\Stoch$.
As a special case of Theorem 5.1 of \cite{cornish2024stochastic}, we then obtain a bijection analogous to \eqref{eq:relative-left-adjoint-3} in this context.
In turn, this yields a recipe for symmetrising Markov kernels analogous to \eqref{eq:sym-characterisation} as follows:
\begin{equation} \label{eq:partial-adjoint-topstoch}
  \begin{tikzcd}[column sep=3.5em]
    \Stoch^H(R X, R Y) \ar{r}{\cong} & \Stoch^G(G/H \otimes X, Y) \ar{rr}{\text{Precompose by \eqref{eq:precomposition-morphism}}} && \Stoch^G(X, Y),
  \end{tikzcd}
\end{equation}
where now $\gamma : X \to G/H$ is a morphism in $\Stoch^G$, hence a $G$-equivariant Markov kernel.
Again, this is the only obvious approach that works without further assumptions.

The symmetrised Markov kernel $\sym(k)$ obtained from \eqref{eq:partial-adjoint-topstoch} has the same form as \eqref{eq:symmetrised-morphism}, with $f$ replaced by $k$. 
However, since the components involved are now Markov kernels, the interpretation is different: given an input $x \in X$, we now \emph{sample} from $\sym(k)(dy|x)$ via 
\[
  \bm{C} \sim \gamma(dc|x) \qquad
  \bm{G} \sim s(dg|\bm{C}) \qquad \bm{Y} \sim k(dy|\bm{G}^{-1} \cdot x) \qquad \text{return $\bm{G} \cdot \bm{Y}$}.
\]
Here $s$ again amounts to a choice of coset representatives \citep[Remark 5.2]{cornish2024stochastic}.
If $G$ is compact, we may take $(s \circ \gamma)(dg|x)$ to be the Haar measure on $G$ \cite[Example 6.1]{cornish2024stochastic}.
When $k$ is an \emph{unconditional} distribution on $Y$, so that $k(dy|x)$ does not depend on $x$, this recovers the symmetrisation approach in Section 4 of \cite{gelberg2024variational}.

\paragraph{Averaging}

The symmetrisation procedure $\sym$ just described produces a Markov kernel $\sym(k)$ that is in general stochastic.
When $Y = \R^d$, we can obtain a deterministic result by computing
\[
  \ave(\sym(k))
\]
where here $\ave$ denotes the \emph{expectation operator} defined as
\[
  \ave(m)(x) \coloneqq \int y \, m(dy|x),
\]
where $x \in X$.
If $G$ acts on $Y$ linearly, then $\ave(\sym(k))$ is also $G$-equivariant \citep[Proposition 4.14]{cornish2024stochastic}.
When $H$ is the trivial subgroup, this recovers the method of \cite{jinwoo2023learning}, and by implication other related methods such as Janossy pooling \citep{murphy2018janossy} and frame averaging \citep{puny2022frame}.

However, $\ave$ can be a costly operation to compute, especially in high dimensions, and may not even be defined if $Y$ is not a convex space like $\R^d$.
By working with the stochastic condition \eqref{eq:stochastic-equivariance} directly, the symmetrisation approach of \cite{cornish2024stochastic} provides a means for avoiding these issues while still obtaining symmetry guarantees from the model overall.

\bibliography{main}

\newpage

\appendix

\section{Basic definitions of category theory} \label{apd:basic-category-theory}

A highly accessible introduction to category theory can be found in \cite{perrone2023starting}.
We provide only the basic definition of a category here, which is all we require in the main text.
A \emph{category} consists of a collection of \emph{objects} and a collection of \emph{morphisms}.
Each morphism $f$ has two associated objects $X$ and $Y$ referred to as its \emph{domain} and \emph{codomain} respectively, which we denote by $f : X \to Y$.
Given morphisms $f : X \to Y$ and $g : Y \to Z$, we can obtain their \emph{composition} $g \circ f$, which is a morphism $X \to Z$.
In addition, for every object $X$, there is an identity morphism $\id_X : X \to X$ which interacts with composition in the obvious way.
Precisely, it holds that $f \circ \id_X = f$ and $\id_X \circ h = h$ for all $f : X \to Y$ and $h : Z \to X$.

\paragraph{Example}

Perhaps the most familiar example of a category is $\Set$, whose objects $X, Y, \ldots$ are sets, and whose morphisms $X \to Y$ are just functions $f : X \to Y$.
We can always compose pairs of functions whose domain and codomain match, and moreover every set admits an identity function, so that $\Set$ is indeed a category.

\section{Symmetrising along a homomorphism} \label{apd:sym-along-homomorphism}

\cite{cornish2024stochastic} formulates symmetrisation procedures in slightly more general terms than \eqref{eq:sym-proc-general-form}.
In particular, their starting point is an arbitrary group homomorphism $\varphi : H \to G$.
As for the subgroup approach described above, any $G$-set $X$ again yields an $H$-set $R_\varphi X$ with the same underlying set as $X$, and its $H$-action defined as
\[
  \alpha_{R_\varphi X}(h, x) \coloneqq \alpha_X(\varphi(h), x).
\]
This always satisfies the axioms of an action since $\varphi$ is a homomorphism.
By Definition 4.7 of \cite{cornish2024stochastic}, a symmetrisation procedure in $\Set$ is then any function of the form
\begin{equation} \label{eq:symmetrisation-along-homomorphism}
  \begin{tikzcd}
    \Set^H(R_\varphi X, R_\varphi Y) \ar{r}{\sym} & \Set^G(X, Y).
  \end{tikzcd}
\end{equation}
The case for a subgroup $H \subseteq G$ from \eqref{eq:sym-proc-general-form} can then be recovered by letting $\varphi$ be the subgroup inclusion $H \hookrightarrow G$.

In practice, most homomorphisms $\varphi$ of interest seem to correspond to a subgroup inclusion in some way, at least in spirit.
However, symmetrisation ``along a homomorphism'' like this can be cleaner to talk about for more complex use-cases.
For example, given groups $H$ and $K$, there is an obvious homomorphism
\[
  H \to K \times H
\]
that sends $h \mapsto (e_K, h)$, where $e_K \in K$ is the identity element.
With \eqref{eq:symmetrisation-along-homomorphism} we can talk about symmetrising along this homomorphism directly, whereas we would otherwise need to define the subgroup
\[
  \{(e_K, h) \mid h \in H \} \subseteq H \times K,
\]
which is of course isomorphic to $H$, but somewhat more unwieldy to write down.

The homomorphism approach is particularly convenient for describing the composition of multiple symmetrisation procedures in sequence (see Section 4.5 of \cite{cornish2024stochastic}).
For example, for 3D point cloud data, it is often of interest to obtain a model that is $S_n \times E(3)$ equivariant, where $S_n$ is the symmetric group and $E(3)$ is the Euclidean group (see e.g.\ \cite[Section 3.2]{jinwoo2023learning}).
One approach here would be to symmetrise in sequence along the obvious homomorphisms
\begin{equation} \label{eq:compositional-symmetrisation-example}
  \begin{tikzcd}
    S_n \ar{r} & O(3) \times S_n \ar{r} & E(3) \times S_n,
  \end{tikzcd}
\end{equation}
where $O(3)$ is the orthogonal group.
This would start with a model that is $S_n$-equivariant (such as a transformer without positional encodings \citep{vaswani2023attentionneed}), then upgrade this to become also $O(3)$-equivariant, and finally upgrade this again to become fully $E(3)$-equivariant.
The diagram \eqref{eq:compositional-symmetrisation-example} summarises this process directly, whereas this again becomes more unwieldy to write down in terms of subgroups.

\section{Proof of Theorem \ref{prop:relative-left-adjoint}} \label{apd:proof-of-theorem}

\begin{proof}
  It follows from the assumption that $f$ is $H$-equivariant that $f^\sharp$ is well-defined, since we have
  \begin{align*}
    f^\sharp([gh], x) &= (gh) \cdot f((gh)^{-1} \cdot x) \\
      &= (gh) \cdot f(h^{-1} \cdot g^{-1} \cdot x) \\
      &= (gh) \cdot h^{-1} \cdot f(g^{-1} \cdot x) \\
      &= g \cdot f(g^{-1} \cdot x) \\
      &= f^\sharp([g], x).
  \end{align*}
  Now, given $f^\sharp$, we can recover $f$ since
  \[
    f^\sharp([e], x) = e \cdot f(e^{-1} \cdot x) = f(x),
  \]
  where $e \in G$ is the identity element.
  This shows \eqref{eq:relative-left-adjoint-3} is injective.
  On the other hand, given any $h : G/H \otimes X \to Y$ in $\Set^G$, letting $f(x) \coloneqq h([e], x)$, we recover $f^\sharp = h$ since
  \begin{align*}
    f^\sharp([g], x)
      &= g \cdot f(g^{-1} \cdot x) \\
      &= g \cdot h([e], g^{-1} \cdot x) \\
      &= h(g \cdot [e], g \cdot g^{-1} \cdot x) \\
      &= h([g], x),
  \end{align*}
  where we use the assumption that $h$ is $G$-equivariant in the third step.
  This shows \eqref{eq:relative-left-adjoint-3} is surjective, hence a bijection.
\end{proof}

\paragraph{Categorical explanation}

This result arises very naturally from the perspective of category theory.
The idea is that $R$ is actually part of a \emph{functor} with a left adjoint $E$ as shown:
\[
  \begin{tikzcd}
    \Set^H \ar[shift left=.5em]{rr}{E} & {\scriptstyle \bot} & \Set^G \ar[shift left=.5em]{ll}{R}.
  \end{tikzcd}
\]
This is a classical result (see e.g.\ \cite[(1.4)]{may1996equivariant}).
In particular, the existence of this adjunction means that
\begin{equation} \label{eq:adjunction-proof-step}
  \Set^H(R X, R Y) \cong \Set^G(E R X, Y).
\end{equation}
It is also classical to show that there is an isomorphism of $G$-sets
\begin{align*}
  E R X &\cong G/H \otimes X
\end{align*}
(see e.g.\ \citep[(1.6)]{may1996equivariant}).
Substituting this into \eqref{eq:adjunction-proof-step} yields the desired bijection
\[
  \Set^H(R X, R Y) \cong \Set^G(G/H \otimes X, Y).
\]
The proof of Theorem \ref{prop:relative-left-adjoint} simply describes in more detail exactly how this bijection can actually be computed.

\section{Technical details around Markov kernels} \label{apd:markov-kernels}

\paragraph{Definition}

Technically, a Markov kernel $k : X \to Y$ is a function of the form
\begin{equation} \label{eq:markov-kernel-uncurried}
  k : \Sigma_Y \times X \to [0, 1],
\end{equation}
where $X$ and $Y$ are measurable spaces, such that the function $x \mapsto k(B|x)$ is measurable for every $B \in \Sigma_Y$, and the function $B \mapsto k(B|x)$ is a probability measure for every $x \in X$.
Here $\Sigma_Y$ denotes the $\sigma$-algebra associated to $Y$.

\paragraph{The category $\Stoch$}

Markov kernels give rise to a category called $\Stoch$.
Formally, this has measurable spaces as its objects.
Given two measurable spaces $X$ and $Y$, the morphisms $X \to Y$ are then just the Markov kernels of this form.
Each identity kernel $\id_X : X \to X$ is obtained as
\[
  \id_X(dy|x) \coloneqq \delta_x(dy),
\]
where the right-hand side denotes the Dirac measure at $x \in X$.
Composition is performed via the \emph{Chapman-Kolmogorov formula}: given Markov kernels $k : X \to Y$ and $m : Y \to Z$, we obtain $m \circ k : X \to Z$ with
\[
  (m \circ k)(B|x) \coloneqq \int m(B|y) \, k(dy|x) 
\]
for all $B \in \Sigma_Z$ and $x \in X$.
Intuitively, to sample from $(m \circ k)(dz|x)$, we just sample from $k$ and $m$ in sequence as follows:
\[
  \bm{Y} \sim k(dy|x) \qquad \bm{Z} \sim m(dz|\bm{Y}) \qquad \text{return $\bm{Z}$.}
\]
For a more detailed overview of $\Stoch$, see e.g.\ Section 4 of \cite{fritz2020synthetic}.

\end{document}